\documentclass[10pt,letterpaper]{article}
\pdfoutput=1
\usepackage{cogsci}

\cogscifinalcopy 

\usepackage{hyperref}
\usepackage{graphicx}
\usepackage{pslatex}
\usepackage{apacite}
\usepackage{enumitem}
\usepackage{amssymb}
\usepackage{bbm}
\usepackage{float} 

\title{A Computational Account Of Self-Supervised Visual Learning From Egocentric Object Play}
 
\author{\large \bf Deepayan Sanyal, Joel Michelson, Yuan Yang, James Ainooson \& Maithilee Kunda \\
Department of Computer Science \\
Vanderbilt University \\
Nashville, TN 37212, USA \\
\{deepayan.sanyal, joel.p.michelson, yuan.yang, james.ainooson, mkunda\}@vanderbilt.edu
}

\begin{document}

\maketitle
\begin{abstract}
Research in child development has shown that embodied experience handling physical objects contributes to many cognitive abilities, including visual learning. One characteristic of such experience is that the learner sees the same object from several different viewpoints. In this paper, we study how learning signals that equate different viewpoints---e.g., assigning similar representations to different views of a single object---can support robust visual learning. We use the Toybox dataset, which contains egocentric videos of humans manipulating different objects, and conduct experiments using a computer vision framework for self-supervised contrastive learning. We find that representations learned by equating different physical viewpoints of an object benefit downstream image classification accuracy. Further experiments show that this performance improvement is robust to variations in the gaps between viewpoints, and that the benefits transfer to several different image classification tasks.  

\end{abstract}

\section{Introduction}

In interacting with the real-world, an individual's experience is highly connected from one instant to the next. 
If someone is holding a spoon at one moment, it is likely that they will still be holding the same spoon in the next, possibly at a slightly different distance and hand/head/spoon pose. This physical continuity serves to generate a multitude of different views of the held object.  Furthermore, the physical act of holding the object informs the learner that the sequence of differing views is tied to the same object, i.e. a form of object permanence.  Even if the observer does not know that an object is a spoon, they understand that the object is the same across multiple moments in time. In this paper, we study whether this embodied experience of seeing different views of an object, and knowing that the views correspond to the same object, can provide a useful form of self-supervisory signal to enable visual learning in computational models.

There is a rich body of research studying the links between motor development in infants and their perceptual and cognitive abilities. \shortciteA{bushnell1993motor} proposed that the progressive development of different motor abilities in infants leads to different schedules for various kinds of perceptual inputs; these, in turn, cause a temporal difference in the appearance of various cognitive abilities. Further studies have elaborated on the links between different kinds of perceptual inputs in development and the appearance of various cognitive skills \shortcite{needhamimprovements, 
libertus2010teach, 
schwarzer2013crawling, baumgartner2011infants}. Looking specifically at the ability to hold and manipulate objects, there is evidence that being able to perform hand-held object manipulations benefits several different cognitive abilities, such as learning nouns 
\shortcite{
slone2019self}, visual understanding \shortcite{ruff1982role, soska2010systems} and understanding causality of actions \shortcite{rakison2012does}. Recent research aiming to characterize infant visual experience using head-mounted cameras has found that first-person visual experience of manipulating objects during self-play constitutes a significant portion of the infants' visual diets \shortcite{herzberg2022infant}. In addition, there is considerable consistency in the distributions of object viewing experience across different cultures \shortcite{casey2022sticks}.

\begin{figure}[t!]
  \centering
  \includegraphics[width=0.9\linewidth]{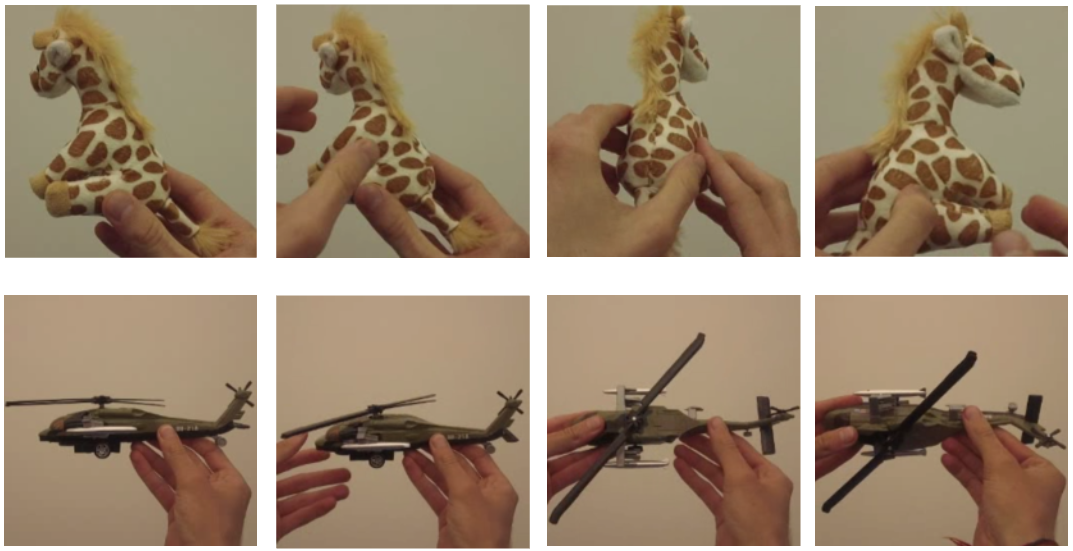}
  \caption{Visual experience during embodied object manipulation. Each row shows frames from an egocentric video of one object being manually rotated. Equating different physical views provides a strong learning signal. }
  \label{fig:toybox-views}
\end{figure}

While previous studies have observed the importance of the embodied experiences generated by infants, the specific learning mechanisms that link these inputs (and their characteristics and distributions) to learning outcomes are not well understood. In this paper, we consider the visual experience that is generated during embodied manipulation of objects and propose a possible mechanism by which this experience helps develop good visual representations which support category learning. To do this, we use the SimCLR framework \shortcite{chen2020simple} which learns effective representations by maximizing the representational similarity between two differently-augmented versions of one image. This framework relies on instance-level similarity to learn representations, and a similar framework has recently been proposed to explain the representational goal of the visual system \shortcite{konkle2022self}. We hypothesize that natural visual experience provides stronger signals for this kind of learning. In this paper, we focus on the multi-view aspect of natural visual experience and show that access to different physical views of the same object leads to emergence of strong category structure. 

Our work is also linked to research showing that temporal contiguity of visual experience can play a crucial role in learning invariant representations \cite{sprekeler2007slowness, li2010unsupervised, wood2018development}. 
Further, the development of such invariant object representations is not affected by reward \cite{li2012neuronal}, suggesting an unsupervised mechanism which regulates this kind of learning. For our part, we only consider the different views of an object that are generated during embodied manipulation of the object and show that equating these views presents a strong signal for category learning. Our contributions in this paper are:
\begin{itemize}[nolistsep, noitemsep]
    \item We demonstrate that representations learned by maximizing similarity between different physical views of the same object support strong performance on a subsequent classification task.
    \item We show that the representations are fairly robust to variations in the magnitude of difference between the paired object views utilized for learning.
    \item We demonstrate that these learned representations also successfully transfer to a diverse set of downstream classification tasks. 
\end{itemize}

\section{Related Work}
There has been recent interest in using machine learning (ML) models to explain and understand different facets of human visual abilities as they relate to human visual experience. \shortciteA{bambach2018toddler} used convolutional neural networks (CNN) to investigate the differences in the visual experiences of infants and adults and showed that an infants' visual experience contains a more diverse range of views of objects, which lends itself to better object recognition performance. \shortciteA{stojanov2019incremental} addressed the problem of learning object representations from incremental experience with individual objects and showed that repeated experiences with objects help ML models avoid problems related to catastrophic forgetting. 

A recent work \shortcite{orhan2020self} considered the problem of learning representations from infant headcamera recordings without explicit image labels. They used data from the SAYCam dataset \shortcite{sullivan2020saycam}, and showed that a learning signal based on temporal continuity enables learning representations that support image classification on the SAYCam and the Toybox datasets. While this work has similarities to our work, we focus on the visual experience that is generated during embodied object manipulation. 

Other works have used CNNs to reason about the relationship between visual abilities in humans and limitations in visual experience; \shortcite{vogelsang2018potential} showed that CNNs can help explain deficits in configural face processing in children born with congenital cataracts. \shortciteA{jang2021convolutional} showed that while CNNs can be used to recreate differences between object and face processing, they do not yet account for robustness of adult vision to image blur.

Another relevant body of research is that of learning representations from visual data without explicit labels in the field of computer vision. Initial approaches for these methods used various pretext tasks such as image colorization \shortcite{zhang2016colorful}, predicting relative patches in images \shortcite{doersch2015unsupervised}, solving jigsaw puzzles \shortcite{noroozi2016unsupervised} and predicting rotations \shortcite{gidaris2018unsupervised} to generate self-supervision. However, a recent body of work \shortcite{grill2020bootstrap, misra2020self, chen2020simple} 
based on contrastive learning \shortcite{hadsell2006dimensionality} has significantly outperformed those earlier approaches. Self-supervised approaches have also been applied to the problem of learning visual representations from videos \shortcite{wang2015unsupervised, wang2020self, qian2021spatiotemporal, tschannen2020self}. 


\section{Our Approach}

\subsection{Dataset}
\begin{figure}[b!]
  \centering
  \includegraphics[width=0.9\linewidth]{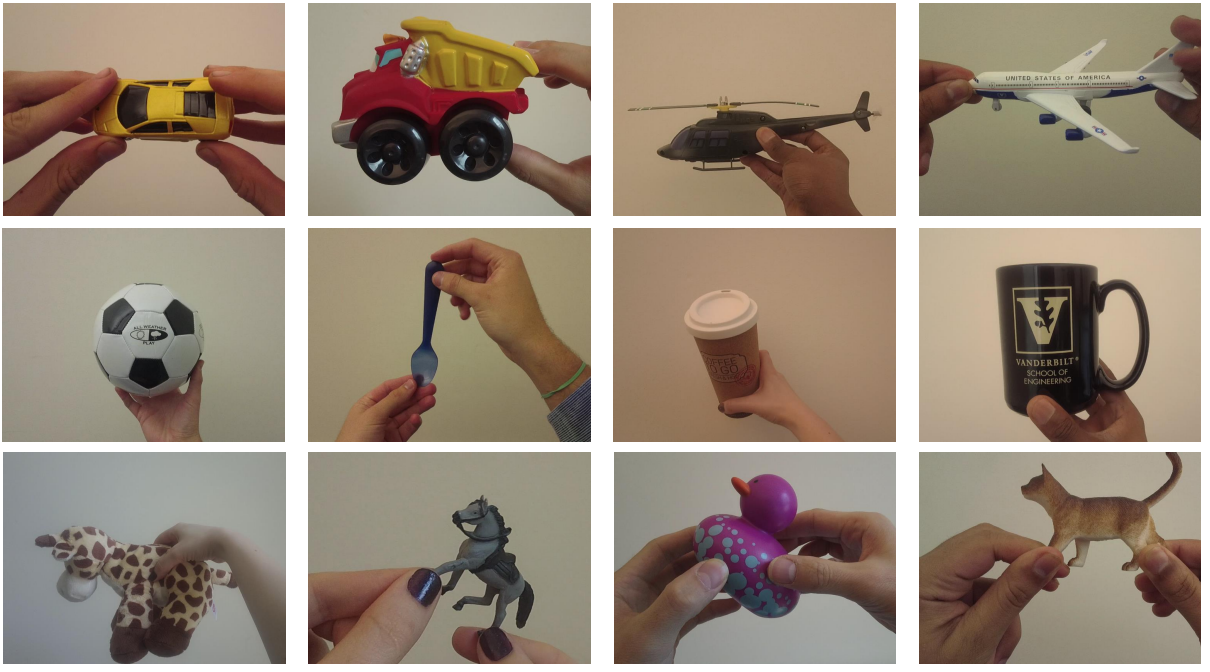}
  \caption{Examples of all 12 classes in the Toybox dataset: car, truck, helicopter, plane, ball, spoon, cup, mug, giraffe, horse, duck, cat.  This figure shows full images; in our experiments, we used images cropped to their bounding boxes.}
  \label{fig:toybox}
\end{figure}

Previous research has established differences between the distributional properties of infant visual experience and traditionally popular datasets used in the computer vision literature \shortcite{smith2017developmental}. Therefore, we used the Toybox \shortcite{wang2018toybox} dataset, which was designed to contain more human-like continuous videos of egocentric handheld object manipulations. The dataset consists of 12 categories from 3 super-categories: household items (ball, cup, mug, spoon), animals (cat, duck, giraffe, horse) and vehicles (airplane, car, helicopter, truck). These 12 categories are among the most common early-learned nouns for children in the US \shortcite{fenson2007macarthur}. For vehicle and animal categories, the objects in the dataset are either realistic, scaled-down models or toy objects. Fig \ref{fig:toybox} shows one object per category from the Toybox dataset.

\begin{figure*}[h!]
  \centering
  \includegraphics[width=0.8\linewidth]{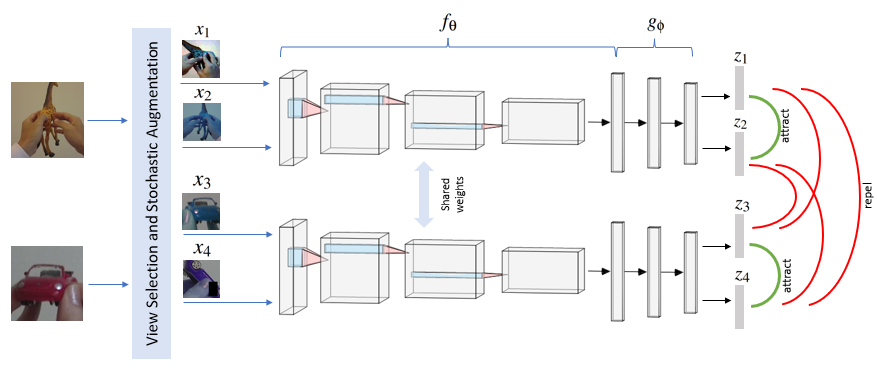}
  \caption{An overview of the learning framework with four images from a batch. Each of the four augmented images are run through the network and we obtain the feature vector $z_i$ associated with each image. The contrastive learning signal then works by moving the positive image pairs closer together while pushing the negative image pairs further apart. The pairs of images linked by the green arcs represent the positive pairs, while the image-pairs linked by the red arcs represent the negative pairs.}
  \label{fig:ssl-approach}
\end{figure*}

The dataset consists of short videos, each of which shows one object being manipulated in one of several ways using an egocentric head-mounted wearable camera. The manipulations present in the dataset include systematic transformations, such as rotation and translation as well as random manipulations labeled as ``hodgepodge'' videos. Since our learning signal uses different viewpoints for each object, we use the 6 rotation videos (one around each axis in one direction) and the hodgepodge video. This gives us a total of 2520 videos for the 360 objects. Each video is about 20 seconds in length, and rotation videos contain two full revolutions around the specified axis and direction.  

There are several interesting aspects of the Toybox dataset.  First, since the objects are being manipulated by hand, the objects are often partially occluded by the subjects' hands. Second, there are several views for each object, including a lot of non-canonical views. Third, unlike traditional ImageNet-style datasets which contain many thousands or millions of objects (with one image each), Toybox has images from a relatively small set of physical instances (30 objects per category) with a large number of images from each.  Thus, it can be challenging for a learner to acquire category-general representations that are less sensitive to the idiosyncrasies of individual objects in the training data. However, these specific aspects of the dataset enable our experiments, since these properties also characterize the visual experience of infants.

Bounding box annotations at 1 fps are available for the rotation and the hodgepodge videos in the Toybox dataset. In order to maintain the original aspect ratios of the objects in the images, we extended the bounding boxes along their shorter dimension to match the size of the larger dimension. Cropping each image to this extended bounding box helps maximize the information content in the images while also preventing distortion of the images.

\subsection{Method}

\subsubsection{SimCLR framework} We use the paradigm of contrastive learning in our experiments, and particularly the SimCLR approach \shortcite{chen2020simple}. The experiments progress in two steps: 

\begin{enumerate}[nolistsep,noitemsep]
\item \textit{Self-supervised representation learning.}  First, a CNN backbone \shortcite{lecun1998gradient} is trained from scratch to learn image representations. During this phase of training, a base network $f_{\theta}$ is attached to a smaller projection network $g_{\phi}$, and this combined network is trained using a self-supervised objective function. 
\item \textit{Representation evaluation using supervised learning.}  In the second phase of training, called the linear evaluation phase, we throw away the projection network $g_{\phi}$, the backbone $f_{\theta}$ is frozen and we attach a linear classifier $fc$ on top of the backbone network. This linear classifier is then trained to perform image classification.
\end{enumerate}

\noindent We now describe the learning signal used for training the network. During training, each minibatch $M$ contains $N$ pairs of images $\{x_{2i}, x_{2i+1}\}_{i=1}^{N}$. Each pair $(x_{2i}, x_{2i+1})$ forms a positive pair and all other image pairs $(x_i, x_k)$ within $M$ constitute the negative pairs. Each image is passed through the backbone and the projection network to obtain $z_i=g\circ f(x_i)$. The loss for one pair of positive images $(x_i, x_j)$ is given by 

\[l(i, j) = - \log \frac{exp(sim(z_i, z_j) / \tau )}{\sum_{k=1}^{2N} \mathbbm{1}_{[k \neq i]} exp(sim (z_i, z_k) / \tau )}\]

where $sim(u, v)$ represents the dot product $u\cdot v$, $\tau$ is the temperature parameter which modulates how sharp the similarity function is and $\mathbbm{1}$ represents the indicator variable, which evaluates to 1 when $k \neq i$ and to $0$ otherwise. The above loss function is called the NT-XEnt loss. For the entire minibatch, the loss function for all positive pairs is aggregated as:

\[\mathcal{L} = \frac{1}{2N} \sum_{k=1}^{N} [l(2k, 2k+1) + l(2k+1, 2k)]\]

By minimizing the above loss function, the learning signal encourages the network to learn representations so that the positive image pairs are closer in the representation space, while the negative image pairs are further away. The effectiveness of the learning signal depends on the positive image pairs that are used. In the original paper \shortcite{chen2020simple}, $x_i$ and $x_j$ are sourced from the same image with different amounts of stochastic image augmentation applied on them, thus telling the network to put differently augmented versions of the same image closer in the feature space compared to different images.

\subsubsection{Modifications and Details} \footnote{The code for these experiments can be found at: \href{https://github.com/aivaslab/toybox_simclr}{https://github.com/aivaslab/toybox\_simclr}} In our experiments, we investigate the extent to which having access to different physical views of the same object contributes to good representations through self-supervision. Thus, in addition to applying stochastic augmentations on the images, we vary the viewpoints from which the positive image pair are chosen. Thus, by equating these two different views, the underlying network learns to bring the representations of these views closer. Fig \ref{fig:ssl-approach} provides an overview of our learning framework.

We use 27 objects from each Toybox class as the training set. During both phases of training, images from these 324 objects are used to train the network. Classification accuracies are reported on images from the remaining 3 objects. During the linear evaluation phase, in keeping with prior work, we use a randomly sampled $10\% $ of the images to train the network. During both phases of training, we apply the following set of augmentations to all training images: color jitter, random grayscale, random crop, and random horizontal flip. No augmentations are applied to the images while calculating the accuracies. For our backbone $f_{\theta}$, we use a ResNet-18 \shortcite{he2016deep} and the projection head $g_{\phi}$ is a 2-layer neural network.
\section{Experiment 1}

As stated above, we vary the viewpoints that comprise each positive pair during training. In doing so, we are signalling that the different views are from the same object. To systematically study how this signal contributes to the learned representations, we use 5 different settings in our experiments:

\begin{figure}
  \centering
  \includegraphics[width=\linewidth]{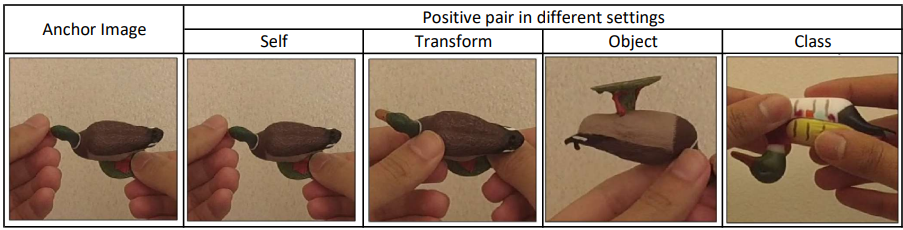}
  \caption{Image pairings used in different experiment settings. In all cases, the anchor image is paired with one other image. In the Self setting, the same image is reused. In the Transform setting, another image of the same object from the same video is selected as the pair. In the Object setting, the image pair can be any image from any of the videos of the object. In the Class setting, the only restriction is that the image pair need to belong to the same class. After the image pair is selected, stochastic image augmentation is applied to both to generate augmented images for learning.}
  \label{fig:ssl-views}
\end{figure}

\begin{enumerate}[nolistsep,noitemsep]
\item SimCLR + Self: The positive pair is sourced from the same image frame with different image augmentations applied. This is the default setting for the SimCLR framework. 
\item SimCLR + Transform: The Toybox dataset consists of 7 videos for each object. In this setting, the positive image pair are sourced from any one of those videos. Specifically, for every image in the dataset, we randomly sample another image from the same video to form the positive pair. 
\item SimCLR + Object: The positive pairs, in this setting, come from any videos of the same object. 
\item Supervised: For baseline comparison, we train a network in a supervised setting on the training images from Toybox. 
\item SimCLR + Class: As a second baseline, we use SimCLR with positive pairs formed by two images from any two objects of the same class. This setting uses the same information about category membership as the Supervised setting but modified to the SimCLR framework.
\end{enumerate}

Fig \ref{fig:ssl-views} shows example image pairings used as positive pairs in these different settings. We observe that the difficulty of the self-supervised task increases from the Self setting to the Class setting as the visual dissimilarity between the positive image pair comes from a larger range.

\begin{table}[ht!]
  \centering
  \includegraphics[width=0.85\linewidth]{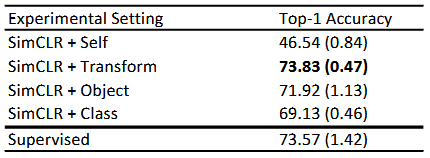}
  \caption{Performance under different training settings.  (Random guessing would yield 1/12, i.e., roughly 8.3\% accuracy.) The best performance is shown by the learner in Transform setting and is comparable to the supervised learner. Accuracy drops off in the Object and Class settings. It is notable that the Transform setting exceeds the performance of the Self setting, which is the default for how SimCLR works. We report the mean and std over two runs with different random seeds. }
  \centering
  \label{fig:simclr_results}
\end{table}

\textbf{Results} Table \ref{fig:simclr_results} shows the results of our experiments in the different settings. We find that the default SimCLR setting achieves modest performance on the Toybox dataset. However, in both the Transform and the Object settings, the final accuracy approaches that of the supervised model. These accuracies show that the representations learned by equating different views of the same object support good classification performance. What we find exciting in the results is that the Transform setting performs so well, despite learning from a weaker supervisory signal compared to the supervised model and the SimCLR models in the Object and Class settings. This seems to suggest that access to some form of viewpoint variation during training is extremely beneficial for the learned representations. We explore this more in Experiment 2.

The model trained in the Class setting did not perform as well as the Transform or the Object settings. This is likely because of the negative pairs: while we control which images form the positive pairs, the negative pairs are automatically decided during training. Because of this, several of the negative pairs are images from the same category. While this drawback is present in the other settings as well, the network seems to be able to handle them better in those settings. This robustness of the learning signal in the Transform and Object settings derives from the fact that in these cases, the chances of getting a negative pair which is more closely related than a positive pair are lower. Hence, the \textit{false negative} pairs do not affect performance in these cases as much.

\section{Experiment 2}

In the previous experiment, we saw that the Transform model performs better than the Object model despite weaker learning signal from the positive pairs. In the current experiment, we wish to study how the visual dissimilarity between the images forming the positive pair affect the learned representations. We do this by carefully controlling the gap between the video frames which form the positive pairs. We focus on the \textit{SimCLR+Transform} configuration in these experiments. We vary the gap between the frames in two settings: 1) Fixed: We fix the gap between the frames, i.e. we say that the two frames forming the positive pair have to be 2 or 4 seconds apart in the same Toybox video. 2) Range: We fix the maximum gap between the two frames , i.e. if we fix the gap to be 2s, the two frames can be 1s or 2s apart. In both settings, we increase the gap in steps of 2s from 0s to 10s and train the networks as described in the previous section. By varying the gap between frames, we can see how the distance in viewpoints for the positive pairs affects the learning performance. It should be noted that a gap of 0 in both settings corresponds to the \textit{SimCLR+Self} model. 

\textbf{Results} Table \ref{tab:gap_1fps} shows our results for these experiments. We see that the \textit{Range} setting seems to perform comparably with the \textit{Fixed}, though it has more variation in the learning signal. This seems to indicate that there is enough variability that arises from the visual data itself which can lead to stronger learning. Further, we see that the performance in both settings remains in the same range even with decreasing gap between the positive pair. 
\begin{table}[ht!]
  \centering
  \includegraphics[width = 0.9\linewidth]{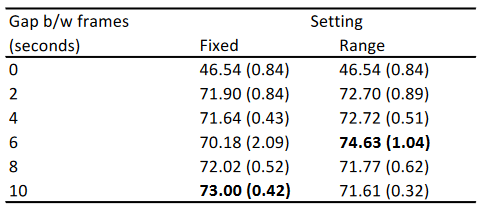}
  \caption{Comparison of model performance using the \textit{SimCLR + Transform} model in the Fixed and Range settings as the gap between frames is varied from 0 to 10 seconds. We report the mean and std over 2 runs.}
  \label{tab:gap_1fps}
\end{table}

\begin{table}[ht!]
  \centering
  \includegraphics[width = 0.9\linewidth]{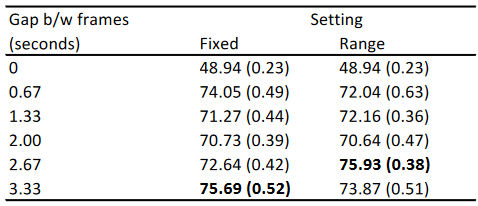}
  \caption{Comparison of model performance using the \textit{SimCLR + Transform} model in the Fixed and Range settings as the gap between frames is varied from 0 to 3.33 seconds. This table uses images from the Toybox dataset extracted at 3fps. We report mean and std over 2 runs.}
  \label{tab:gap_3fps}
\end{table}

To reduce the gap further, we used a version of the Toybox dataset sampled at 3fps. Since the bounding box annotations are done at 1fps, we use linear interpolation to obtain the annotations for the intermediate frames. Further, for this set of experiments, we used only the rotation videos. This allows us to avoid the randomness from the hodgepodge video and study the effect of viewpoint variation in a more structured and regular manner. We increase the gap parameter from 0s to 3.33s in steps of 0.67s. The other settings remain similar to the 1fps experiments. Table \ref{tab:gap_3fps} shows our results in this setting. The first thing we note is that, because the total amount of training data increases close to 3-fold, the accuracy increases in both the \textit{Self} and \textit{Transform} settings. This is consistent with previous results in the machine learning literature showing that more data is generally beneficial. Secondly, we also note that the performance in both settings remains competitive even when the gap between frames is reduced to $0.67$s. This demonstrates that the learning signal remains robust even when the gap between frames is reduced to 0.67s. With the Toybox videos, this gap corresponds to an average angular distance of $12^{\circ}$ between viewpoints. These results suggest that during object manipulation, it is possible to leverage even small variations in viewing angles to learn good visual representations.

\begin{table*}[t!]
  \centering
  \includegraphics[width = 0.75\linewidth]{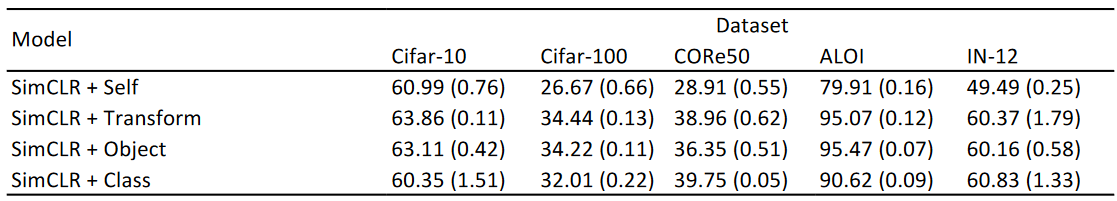}
  \caption{Performance of the models trained with different learning signals on various transfer experiments }
  \label{tab:simclr-transfer}
\end{table*}

\section{Experiment 3}
In the previous experiments, we have seen that representations learned using self-supervision are beneficial for category learning on the Toybox dataset. In the final set of experiments, we examine how these representations generalize to other kinds of classification tasks. Do the benefits we see by equating different physical views of objects in classifying Toybox images transfer to other datasets as well? To accommodate a variety of classification tasks, we use several downstream tasks to measure transfer performance. The phenomenon of machine learning methods developing bias towards their training dataset is well-documented \shortcite{torralba2011unbiased}. Our aim in this set of experiments is to show that the benefit from using the learning signal is not limited only to the Toybox dataset, but extends to other datasets as well. We will refrain from providing a detailed description of the datasets, but will point out some aspects of the datasets which we find relevant for this paper.

In the computer vision community, use of large-scale datasets is mainstream. These datasets function as good test data to evaluate the generality of models. To include these kinds of datasets, we choose the CIFAR-10 and CIFAR-100 datasets \shortcite{krizhevsky2009learning}. While the CIFAR-10 dataset has some classes overlapping with the Toybox dataset, the CIFAR-100 dataset has classes of natural scenes and a much larger variety of classes than the Toybox dataset. While these internet-based datasets have a large number of instances for each class, there is usually only one image of each instance and it has been shown that the images in the dataset have a skewed distribution over viewpoints due to cameraman bias. To include more datasets where evaluation is done over multiple viewpoints, we include an object classification task on the CORe50 \shortcite{lomonaco2017core50} dataset and an instance classification task on the ALOI \shortcite{geusebroek2005amsterdam} dataset.
Finally, we examine if the representations learned from the Toybox dataset are transferable to real-world instances of the same categories. For this, we have curated the IN-12 dataset using images from the popular ImageNet \shortcite{deng2009imagenet} and MS-COCO \shortcite{lin2014coco} datasets for the Toybox classes. Specifically, we identify classes in the ImageNet dataset which overlap with the Toybox classes and randomly sample from each of these candidate classes to select 1700 images for each Toybox category. From these 1700 images, we use 1600 images per class for training and 100 images per class for testing the network.

\subsection{Results}
Table \ref{tab:simclr-transfer} shows our results for the transfer learning experiments. We see that the \textit{Transform} model performs better than the \textit{Self} model and is competitive with the \textit{Object} models on all the transfer tasks. The improvement in performance is strong for the datasets with multiple viewpoints (CORe50 and ALOI), thus showing that learning from multi-view egocentric experience of object manipulation benefits downstream performance for other multi-view datasets as well. The relative jump in performance is highest for CIFAR-100, thus demonstrating the general strength of the learned representations even for classification tasks where the image classes are vastly different. Looking at how the representations learned from the Toybox images transfer to real-world images from the same categories (IN-12 dataset), we find that similar trends hold in this case as well. It is interesting that even in these transfer conditions, the \textit{Class} models generally perform worse than the \textit{Transform} models, though it performs slightly better for the CORe50 dataset.

\section{Conclusion and Discussion}
We have considered the problem of learning from the visual experience of embodied object manipulation and proposed a mechanism by which good representations which support image classification can be learned. We do this by utilizing a learning signal which minimizes the representational distance between different physical views of the same object. Through our experiments, we showed that this signal enables learning good representations which support categorization. We further showed that this signal is robust to the magnitude of difference between the viewpoint-pair which generate the learning signal. Finally, we demonstrated that the generality of learning with this signal by showing that the learned model can transfer non-trivially to a diverse classification tasks.

Our work leads to several important questions that will be addressed in future work: 1) While our work shows the effectiveness of the learning signal for downstream classification tasks, research has shown that similar algorithms can lead to relevant information being lost in the model \shortcite{xiao2021should}. In order to understand the development of robust human vision that can perform diverse visual tasks, further research looking at the interaction between learning signals and the efficacy of the learned representations at different tasks needs to be done. 2) Our approach requires the use of strong image augmentations. This is likely due to the fact that CNNs can learn to use color histograms as a shortcut \shortcite{chen2020simple} during the self-supervised training and this problem is especially acute in the case of exemplar-based datasets like the Toybox dataset. Further research needs to be done to understand how the human visual system avoids such issues.

\section{Acknowledgements}
We would like to thank the anonymous reviewers for their helpful and constructive comments.

\bibliographystyle{apacite}

\setlength{\bibleftmargin}{.125in}
\setlength{\bibindent}{-\bibleftmargin}

\bibliography{references}

\end{document}